\documentclass[runningheads,a4paper]{llncs}

\usepackage{amssymb}
\usepackage{graphicx}
\usepackage{url}
\usepackage{graphicx}
\usepackage{mathtools}
\graphicspath{{./pic/}}

\urldef{\mailsa}\path|{yang.fan.xv6,|
\urldef{\mailsb}\path|jaymar.soriano.jk2,|
\urldef{\mailsc}\path|takatomi-k,|
\urldef{\mailsd}\path|kazushi } @is.naist.jp|    
\newcommand{\keywords}[1]{\par\addvspace\baselineskip
\noindent\keywordname\enspace\ignorespaces#1}

\begin{document}

\mainmatter  

\title{A Hierarchical Mixture Density Network}


\author{Fan Yang, Jaymar Soriano, Takatomi Kubo
\and Kazushi Ikeda}
%

\institute{Graduate School of Information Science, Nara Institute of Science and Technology,\\
Ikoma, Nara, Japan\\
\mailsa~
\mailsb~
\mailsc~
\mailsd\\
}

\toctitle{}
\tocauthor{}
\maketitle

\begin{abstract}
The relationship among three correlated variables could be very sophisticated, as a result, we may not be able to find their hidden causality and model their relationship explicitly. However, we still can make our best guess for possible mappings among these variables, based on the observed relationship. One of the complicated relationships among three correlated variables could be a two-layer hierarchical many-to-many mapping. In this paper, we proposed a Hierarchical Mixture Density Network (HMDN) to model the two-layer hierarchical many-to-many mapping. We apply HMDN on an indoor positioning problem and show its benefit.

\keywords{Mixture Density Network, Hierarchical Many-to-many Mappings}
\end{abstract}

\section{Introduction}

In real problems, it is common to find that the same variable could be generated under different conditions, resulting in different values. Different variables, conversely, could have the same value. Such a relationship between two variables can be denoted as a many-to-many mapping. Although it is difficult to model a many-to-many mapping directly in the continuous space, it can be simplified to a one-to-many mapping when conditioning on one variable. 

\begin{figure}
\centering
\includegraphics[height=3.5cm]{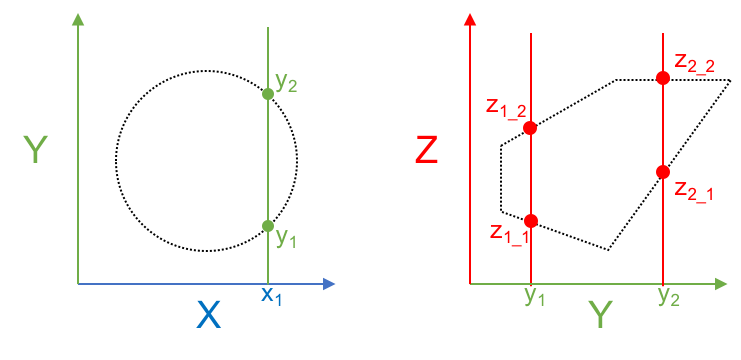}
\vspace*{-0.5cm}
\caption{A hierarchical many-to-many mappings. $x_1$ can be mapped to $y_1$ and $y_2$, while $y_1$ can be mapped to $z_{1\_1}$ and $z_{1\_2}$, and $y_2$ can be mapped to $z_{2\_1}$ and $z_{2\_2}$. If $x_1$ and $z_{1\_1}$ are observed features, the target is most likely to be $y_1$}
\label{m_m_mapping}
\end{figure}

Generally, Mixture Density Network (MDN) is used for one-to-many mapping in a continuous space, as its output is a multi-modal distribution, which is suitable to approximate multiple targets~\cite{bishop1994mixture}. Many applications have indicated the successes of applying MDN to model one-to-many mapping between two variables. Inspired by these works, we consider whether MDN could be used for modeling a two-layer hierarchical many-to-many mapping among three variables (see Fig.\ref{m_m_mapping}).

In this paper, we suppose a two-layer hierarchical many-to-many mapping can be approximated by two many-to-many mappings, connected by a sampling method. Therefore, we proposed a Hierarchical Mixture Density Network (HMDN), by integrating two pre-trained MDNs together. In the indoor positioning experiment, we show that our HMDN can take both WLAN fingerprint and illumination intensity into account directly, and make better predictions. 

\section{Mixture Density Network}

Typical neural networks only consider an unimodal distribution for one output. However, it may not be sufficient to represent the statistics of a complex output. Mixture Density Network (MDN) is a neural network that was designed to model an arbitrary distribution of output, by replacing the unimodal distribution to a linear combination of kernel functions \cite{bishop1994mixture}. In other words, MDN still models the mapping between the input $x$ and the output $y$, but instead of directly giving $y$, it provides the parameters of a mixture density from which we can sample $y$.

For most applications, the kernel function is simply chosen as Gaussian, whose probability density is represented in the form
\begin{equation}
	p (y~|~x;w) = \sum^{K}_{k=1}\pi_{k}(x;w)\mathcal{N}\Big(\mu_{k}(x;w), \sigma^{2}_{k}(x;w)\Big),
\end{equation}with the constraints\begin{equation}
	\sum_{k=1}^{K}\pi _{k}(x) = 1,  ~~0\leq \pi_{k}(x;w)\leq1,
\end{equation}
where K is the number of mixture components, $w$ is the neural network parameters,  $\pi_{k}(x;w)$, $\mu_{k}(x;w)$ and $\sigma_{k}(x;w)$ are the mixing coefficients, the means, and the variances of Gaussian mixtures. 

\begin{figure}
\centering
\includegraphics[height=5cm]{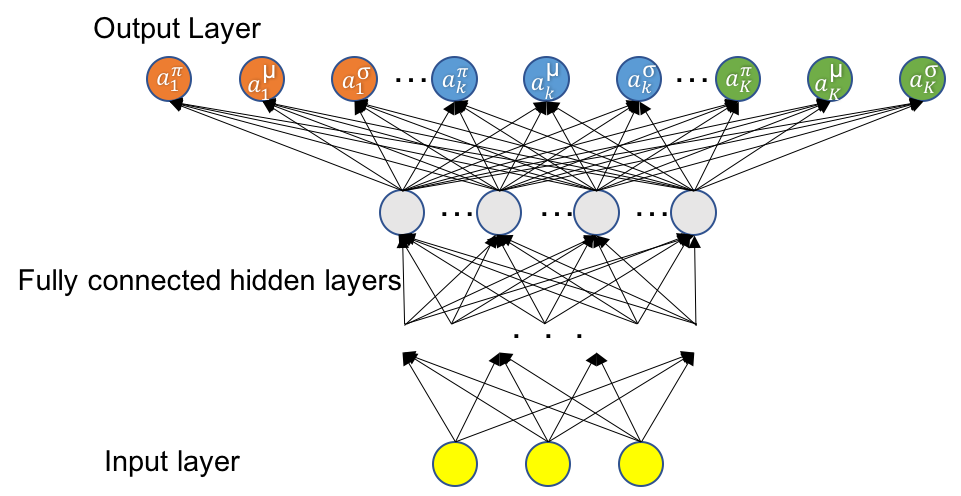}
\caption{Representation of the Mixture Density Network}
\vspace*{-0.5cm}
\label{MDN_graph}
\end{figure}

What MDN output layer generates are activation units, which can be divided into three types as  $a^{\pi}$, $a^{\sigma}$ and $a^{\mu}$ (see Fig.\ref{MDN_graph}). Hence, $\pi_{k}(x;w)$, $\mu_{k}(x;w)$ and $\sigma_{k}(x;w)$ can be derived as follows:
\begin{equation}
\begin{split}
	& \pi_{k}(x) = \frac{exp(a^{\pi}_{k})}{\sum_{l=1}^{K}exp(a^{\pi}_{l})}, \\
    & \sigma_{k}(x) = exp(a_{k}^{\sigma}),\\
    & \mu_{ki}(x) = a_{ki}^{\mu},\\
    & k \in \{1,2,...,K\}, \\
    & i \in \{1,2,...,D\},
\end{split}
\end{equation}
where $D$ is the dimension of $y$. Hence, the number of elements for $a^{\pi}$, $a^{\sigma}$ and $a^{\mu}$ are $K$, $K$ and $D \times K$, respectively.

In MDN, what we want to get is the conditional density function $p(y~|~x)$, for each pair $(x,y)$. Therefore, a MDN is trained by maximizing the conditional density likelihood over all variables,  yielding the following cost function
\begin{equation}
E(w) = -\sum_{n=1}^{N} ln \bigg \{ \sum_{k=1}^{K} \pi_{k}(x_n;w) \mathcal{N}\big(y_{n}~|~\mu_k(x_n;w),\sigma^{2}_{k}(x_n;w) \big) \bigg\}
\end{equation}
Since the derivatives of $E(w)$ and output activations (i.e. $a^{\pi}$, $a^{\sigma}$ and $a^{\mu}$ ) can be calculated~\cite{bishop1994mixture}, MDN training can be performed using general gradient descent algorithms.

\section{Extensions of MDN}

In this section, we review several neural network structures integrated with MDN. Once we get the idea about how they are constructed, HMDN can be treated as a similar construction.

Although MDN is popularly used in a range of applications, there is little difference in the structure of how it integrates with other networks. Since it is difficult to back-propagate the gradient from other networks to an MDN, the common integration is done by putting MDN at the final output layer, modeling the one-to-many mapping between extracted features and final targets. 

\begin{figure}[!htb]
\centering
\includegraphics[height=3.5cm]{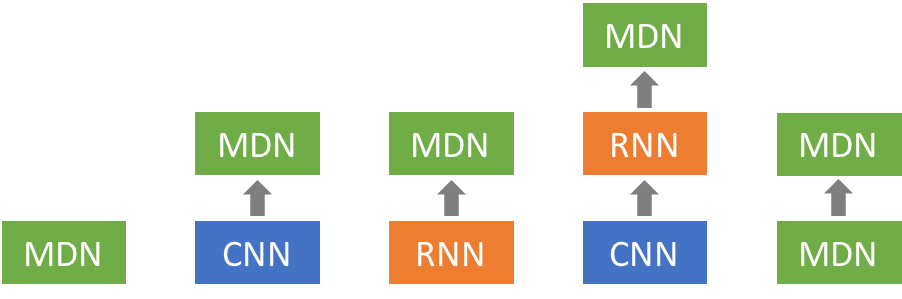}
\vspace*{-0.2cm}
\caption{Extensions of MDN. A single MDN was used in \cite{richmond2006trajectory},\cite{herzallah2004mixture}. CNN+MDN is used in \cite{moon2016predicting}, \cite{iso2017density}. CNN+RNN is used in \cite{wang2016gating},\cite{berio2016sequence},\cite{rehder2017pedestrian}. CNN+RNN+MDN is used in \cite{bazzani2016recurrent}. MDN+MDN is similar to \cite{kingma2013auto}.}
\label{MDN_zoo}
\end{figure}

Denoting the neural networks before MDN as $g_{1}$, while MDN as $g_{2}$,  and their corresponding parameters are $w_1$ and $w_2$, respectively, we can represent aforementioned structure in a general form:
\begin{equation}
\begin{split}
    &g_{1} = f (x;w_1), \\
    &g_{2} = \sum^{K}_{k=1}\pi_{k}(g_{1};w_2)\mathcal{N}\Big(\mu_{k}(g_{1};w_2), \sigma^{2}_{k}(g_{1};w_2)\Big), 
\end{split}    
\end{equation}where $f$ could be a convolutional neural network (CNN)~\cite{krizhevsky2012imagenet}, a recurrent neural network (RNN)~\cite{hochreiter1997long}, or a combination of CNN and RNN. To some extent, an variational autoencoder~\cite{kingma2013auto} can be seen as putting two MDNs together symmetrically, in which the mixture density layer is shared. Extensions of MDN are summarized in Fig. \ref{MDN_zoo}.

\section{Hierarchical Mixture Density Network}

\begin{figure}[!htb]
\centering
\includegraphics[height=7cm]{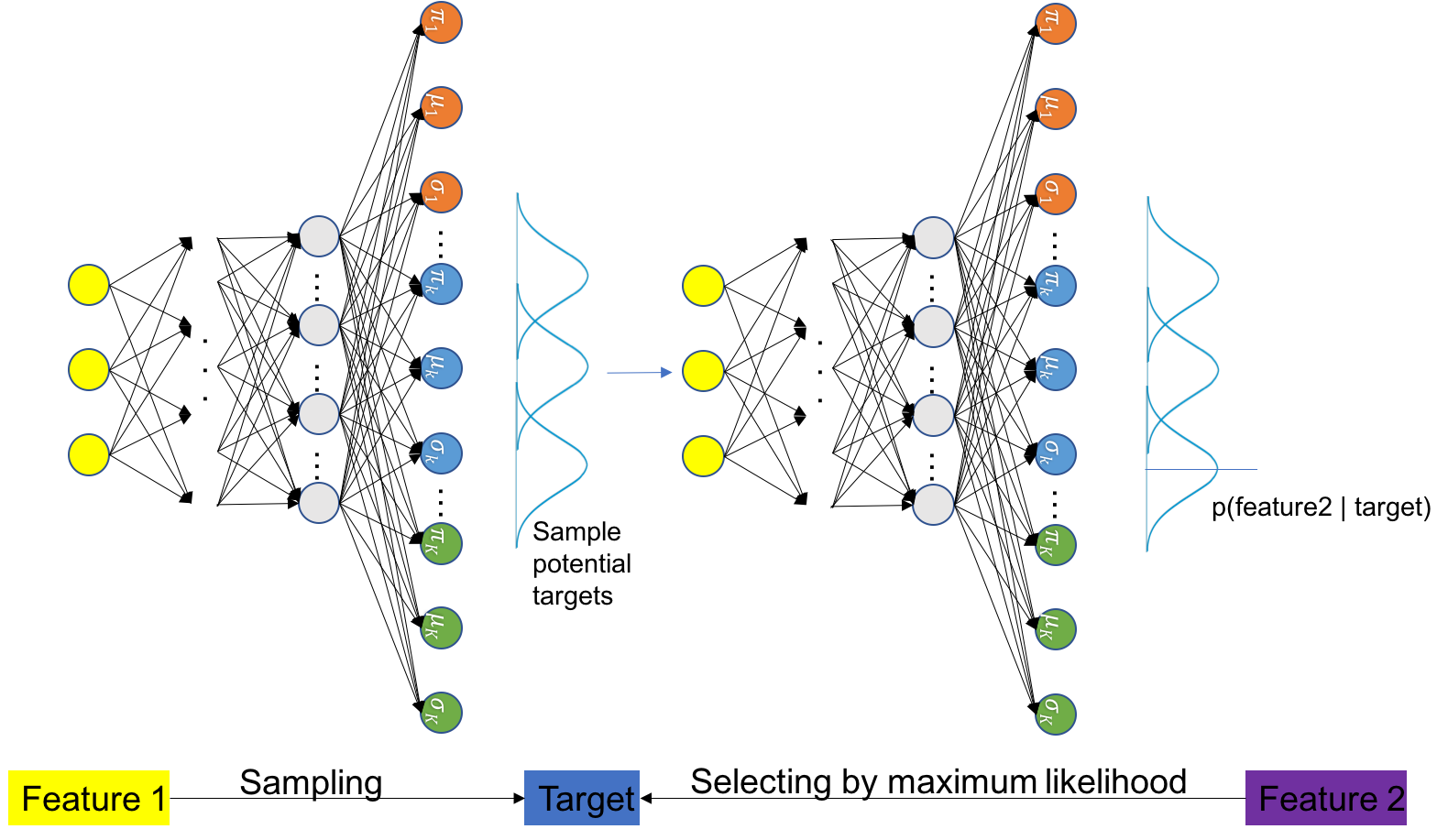}
\vspace*{-0.2cm}
\caption{The model structure of HMDN. Sampling potential targets using feature 1, and further selecting targets using feature 2.}
\label{HMDN_model}
\end{figure}

Referring to Fig.\ref{MDN_zoo}, our proposed HMDN structure can be interpreted as MDN + MDN. Nevertheless, in contrast to a variational autoencoder, the output of the first MDN is the input of the second MDN in HMDN (see Fig.\ref{HMDN_model}). To concrete the idea about HMDN, following example is used.

Suppose we have datasets  $X$ and  $Z$ as features, while $Y$ are the targets. Herein, $Y $ is the bridge to link $X$ and $Z$ together. In general,  we train a model using both $X$ and $Z$ as inputs and $Y$ as the output. However, it may not work when the relations from $X$ to $Y$ and $Y$ to $Z$ are both many-to-many mappings in the continuous space. 

To tackle this issue, we use one MDN ($g_{1}$) to model the mapping from $X$ to $Y$, conditioning on each variable $x$;  while using another MDN ($g_{2}$) to model the mapping from $Y$ to $Z$, conditioning on each variable $y$. The formulas can be represented as
\begin{equation}
\begin{split}
	&g_1 = p(y~|~x;w_1) = \sum^{K}_{k=1}\pi_{k}(x;w_1)\mathcal{N}\Big(\mu_{k}(x;w_1), \sigma^{2}_{k}(x;w_1)\Big),  \\
	&g_2 = p(z~|~y;w_2) = \sum^{K}_{k=1}\pi_{k}(y;w_2)\mathcal{N}\Big(\mu_{k}(y;w_2), \sigma^{2}_{k}(y;w_2)\Big).
\end{split}    
\end{equation}

Given $X$, we can sample $Y_{sample} = \{y_1,y_2,...,y_m\}$ from $g_1(X)$.  We further select samples from $Y_{sample}$, depending on which sample can give higher $p(z~|~y)$. Despite $p(y~|~z)$ is what we truly aim at, based on following Bayesian formula, $p(y~|~z)$ is proportional to $p(z~|~y)$. 
\begin{equation}
\begin{split}
& p(y~|~z)= \frac{p(z~|~y)p(y)}{p(z)},\\
& p(y~|~z)\propto p(z~|~y).
\end{split}
\end{equation}
Since $Z$ are given, after $Y_{sample}$ were sampled,  both $p(y)$  and  $p(z)$ are constants in above formula.

\section{Experiment}

To apply HMDN in a real situation, we utilize
the UJIIndoorLoc Data Set~\cite{torres2014ujiindoorloc}, whose features are WLAN intensity received from mobile phones, while the targets are the two-dimensional coordinates with respect to each WLAN fingerprint. Due to the signal reflection or limited Wireless Access Points (WAPs), the same WLAN fingerprint could be mapped to several potential positions. Conversely, one position could have several fingerprints at the different time, as the signal is unstable. 

In order to make indoor positioning more accurate, besides WLAN fingerprint, we hope to use other location related signals. Since illumination intensity could be location dependent indoor, it may support the indoor positioning~\cite{randall2007luxtrace}. However, even in the same room, illumination intensity does not remain constant. For instance, the illumination intensity could vary in three conditions: sunny outside, cloudy outside and night with light opening. Therefore, using the light meter at a fix position, the illumination intensity could be three levels in the previous assumption. Moreover, several positions could have the same illumination intensity.

Putting WLAN fingerprint and illumination intensity together, there are two many-to-many mappings. Our proposed HMDN is pertinent in such case. 

In the UJIIndoorLoc Data Set, there is no illumination information. For this reason, we customize this dataset by adding simulated illumination intensity under three aforementioned conditions. As our main focus is for demonstration, a simple location dependent luminous field was utilized. Through inspecting the coordinate data of one room in UJIIndoorLoc Data Set, we assume the size of a room could be $17 \times 10$ meters, the average height of holding a phone could be 1.5 meters, and ceiling height could be 4 meters. Without considering any light reflection in the room, and simply treat the sunlight from the window as a point source light, we are able to calculate the illumination intensity at 100 randomly chosen points. In the real scenario, the illumination conditions could be more sophisticated and diverse, but MDN still can adapt to it by adding layers and mixture components. 

Now, we have two observed features as WLAN fingerprints and illumination intensities, and our targets are corresponding coordinates. We first use WLAN fingerprints and associated coordinates to train an MDN denoted by $g_{1}$. Next, we train another MDN denoted by $g_{2}$, modeling the mapping from coordinates to corresponding illumination intensity. Given one WLAN fingerprint $s$, we can sample 100 positions as $A = \{pos_1,pos_2,...,pos_{100}\}$ from $g_{1}(s)$ and calculate the conditional likelihood of illumination intensity $I$ given $A$, i.e., $p( I~|~g_{2}(A))$. Finally, selecting 20 points from maximum  $p( I~|~g_{2}(A))$, corresponding coordinates can be approximated by their mean.

\begin{figure}[h!]
  \begin{tabular}{ccc}
     \begin{minipage}[b]{0.5\hsize}
       \begin{center}
         \includegraphics[clip,height=4.2cm]{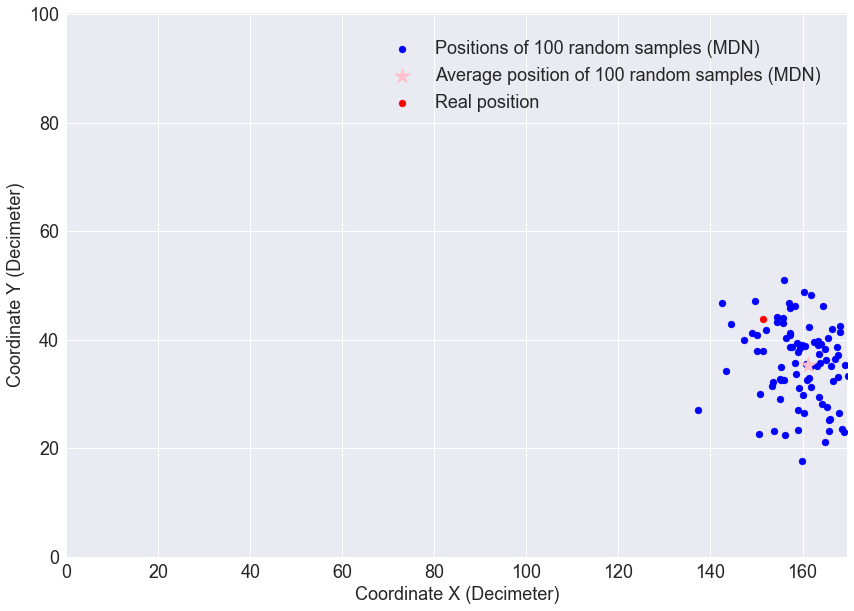}\\
         \vspace*{-0.12 cm}   (a)
       \end{center}
     \end{minipage}
     \begin{minipage}[b]{0.5\hsize}
       \begin{center}
         \includegraphics[clip,height=4.2cm]{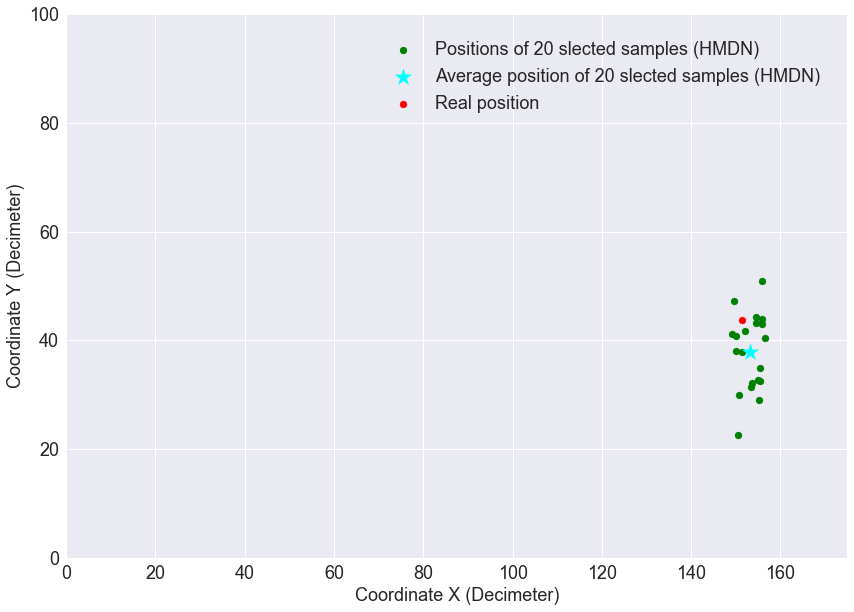}\\
         \vspace*{-0.12 cm}   (b)  
       \end{center}
     \end{minipage}
     
     \\    
   \begin{minipage}[b]{0.5\hsize}
       \begin{center}
         \includegraphics[clip,height=4.2cm]{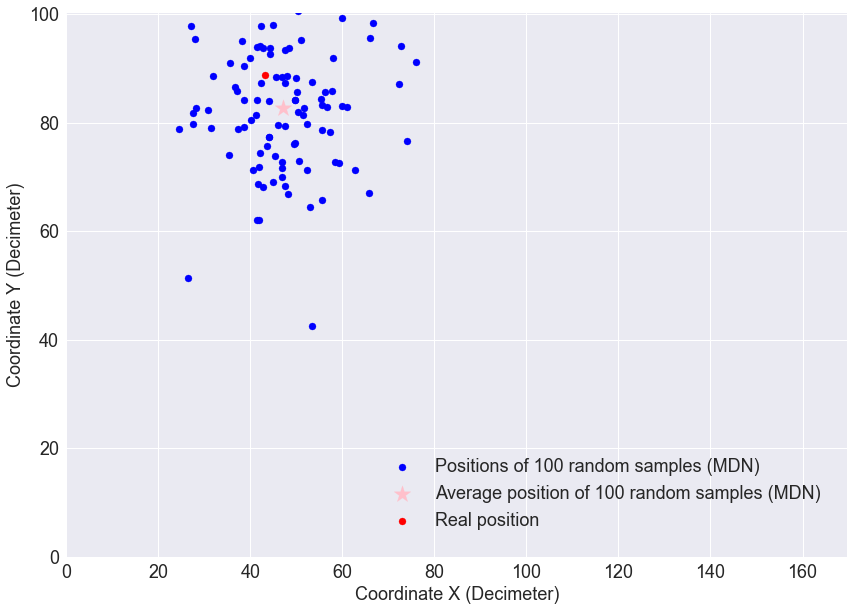}\\
         \vspace*{-0.12 cm}  (c) 
       \end{center}
      \end{minipage}
     \begin{minipage}[b]{0.5\hsize}
       \begin{center}
         \includegraphics[clip,height=4.2cm]{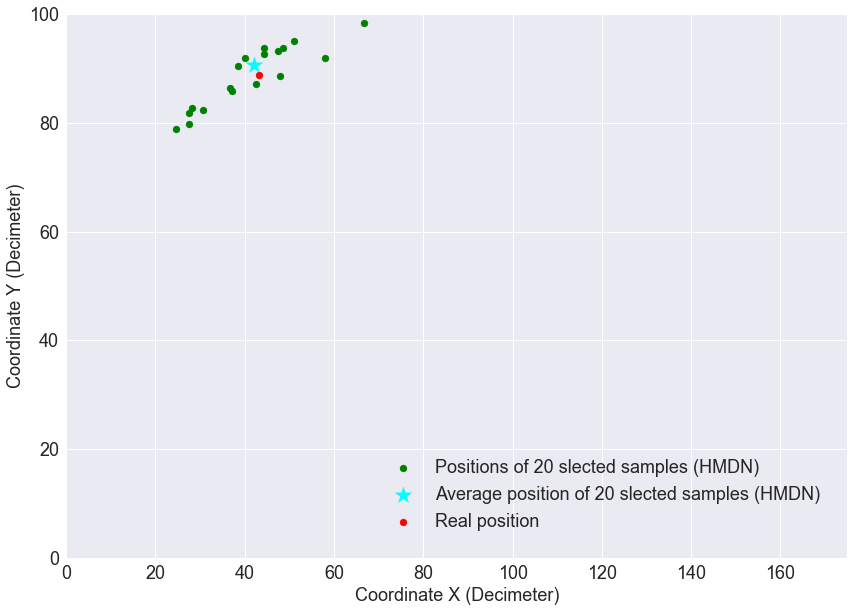}\\
         \vspace*{-0.12 cm}   (d) 
       \end{center}
     \end{minipage}

     \\    
     \begin{minipage}[b]{0.5\hsize}
       \begin{center}
         \includegraphics[clip,height=4.2cm]{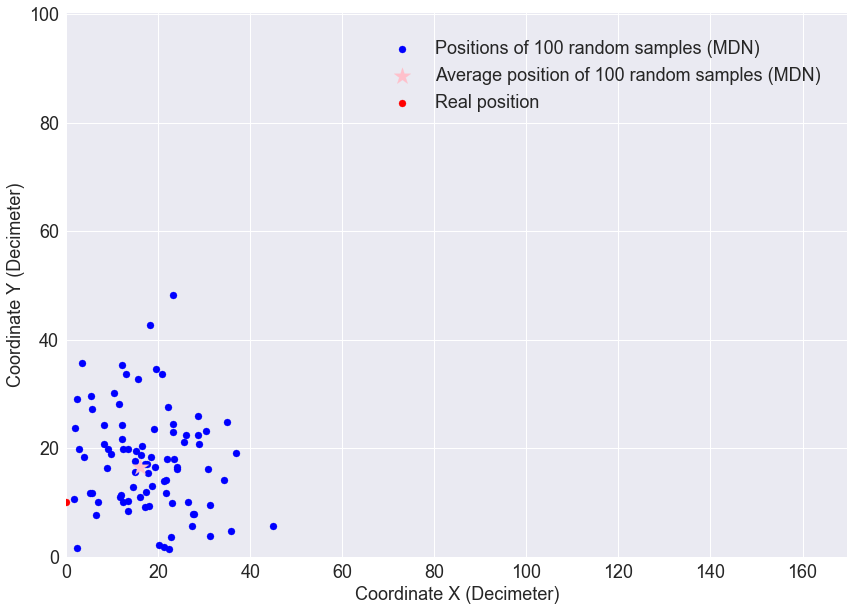}\\
         \vspace*{-0.12 cm}  (e) 
       \end{center}
      \end{minipage}    
     \begin{minipage}[b]{0.5\hsize}
       \begin{center}
         \includegraphics[clip,height=4.2cm]{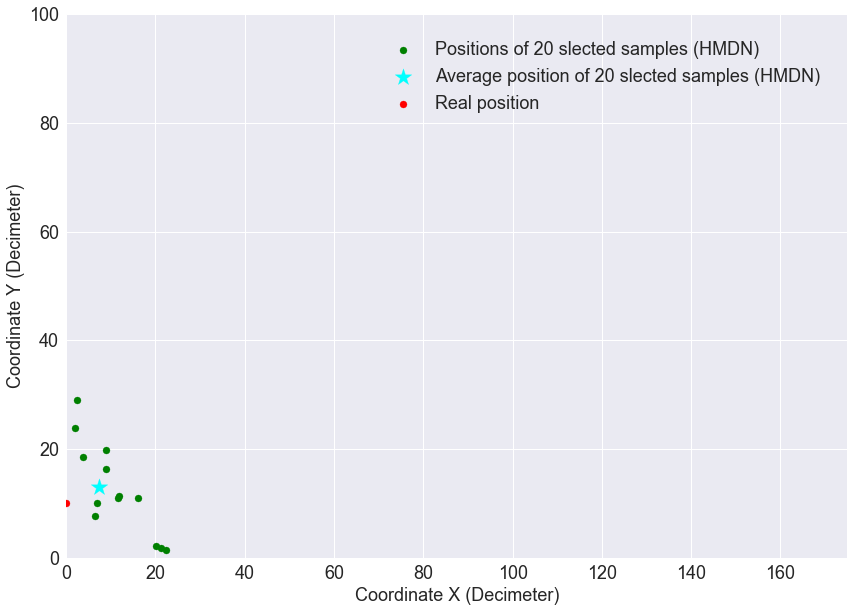}\\
        \vspace*{-0.12 cm} (f)
       \end{center}
     \end{minipage}  
     
       \end{tabular}    
       \vspace*{-0.3cm}
   \caption{Predicted indoor locations. The blue points in (a), (c) and (e) are predicted locations that only depend on WLAN fingerprints. Each of them contains 100 predicted locations. The green points in (b), (d) and (f) are prediction locations that are further selected using illumination intensities. Each of them contains 20 predicted locations.}
   \label{position_test}
\end{figure}

Here, we choose three estimated samples to be shown in Fig.\ref{position_test}.  The coordinates estimation is more accurate by using HMDN to integrate illumination intensity. Even though it is very obvious that results will be improved by adding illumination intensity, what we want to highlight is HMDN can learn such a sophisticated relationship easily while other models cannot.

\section{Conclusion}

To model a hierarchical many-to-many mapping relationship among three variables in a continuous space, we proposed a Hierarchical Mixture Density Network (HMDN), which to the best of our knowledge has not been shown before. We performed an experiment on an indoor positioning problem, where our HMDN can directly model the relationship among position coordinates, WLAN fingerprints and illumination intensities under different conditions. The accuracy of prediction was improved by using HMDN to further consider illumination intensity, which means that HMDN can correctly learn the relationship among three variables.

\bibliographystyle{splncs03}
\bibliography{refs}

\end{document}